\documentclass[11pt,a4paper]{article}
\usepackage[hyperref]{acl2020}
\usepackage{times}
\usepackage{latexsym}

\usepackage{soul}
\usepackage{url}
\usepackage[utf8]{inputenc}
\usepackage{graphicx}
\usepackage{amsmath}
\usepackage{amsthm}
\usepackage{booktabs}
\usepackage{algpseudocode}
\usepackage{array,multirow}
\newcolumntype{?}{!{\vrule width 1.2pt}}
\usepackage{xcolor}
\usepackage{colortbl} 
\usepackage{amssymb}
\usepackage{wrapfig}
\usepackage{tabularx}
\usepackage{microtype}      
\usepackage{subcaption}
\usepackage{xfrac}
\usepackage{mathtools}
\usepackage{todonotes}

\makeatletter
\if@todonotes@disabled

\else

\fi
\makeatother

\usepackage{defs}

\usepackage[vlined,ruled]{algorithm2e}
\SetKwProg{Init}{init}{}{}
\usepackage{nicefrac}       

\usepackage{microtype}

\aclfinalcopy 

\title{$k$-Neighbor Based Curriculum Sampling for Sequence Prediction}

\author{James O' Neill and Danushka Bollegala \\
Department of Computer Science \\
University of Liverpool\\
Liverpool, United Kingdom L69 3BX\\ 
\texttt{\{james.o-neill,danushka.bollegala\}@liverpool.ac.uk} \\}

\date{}

\begin{document}
\maketitle

\begin{abstract}
Multi-step ahead prediction in language models is challenging due to the discrepancy between training and test time processes. At test time, a sequence predictor is required to make predictions given past predictions as the input, instead of the past targets that are provided during training. This difference, known as exposure bias, can lead to the compounding of errors along a generated sequence at test time.
To improve generalization in neural language models and address compounding errors, we propose \textit{Nearest-Neighbor Replacement Sampling} -- a curriculum learning-based method that gradually changes an initially deterministic teacher policy to a stochastic policy. A token at a given time-step is replaced with a sampled nearest neighbor of the past target with a truncated probability proportional to the cosine similarity between the original word and its top $k$ most similar words. This allows the learner to explore alternatives when the current policy provided by the teacher is sub-optimal or difficult to learn from.
The proposed method is straightforward, online and requires little additional memory requirements. We report our findings on two language modelling benchmarks and find that the proposed method further improves performance when used in conjunction with scheduled sampling.
\end{abstract}

\section{Introduction}
Language modeling is a foundational challenge in natural language processing (NLP) and is also important for many related sequential tasks such as machine translation, speech recognition, image captioning and question answering ~\cite{mikolov2010recurrent,sutskever2014sequence}. 
It involves predicting the next token given past tokens in a sequence using a parametric model, $f_{\theta}(\cdot)$, parameterized by $\theta$. This is distinctly different from the standard supervised learning that assumes the input distribution to be i.i.d., which leads to a non-convex objective even when the loss is convex given the inputs.

In supervised learning, the standard modeling procedure\footnote{In the context of RL we will refer to tokens $x$ as actions $a$, models $f$ as policies $\pi$ and predictions $\hat{y}$ as $\pi(s'|a,s)$.} involves training a policy $\hat{\pi}$ to perform actions given an \textit{expert} policy $\pi^{*}$ at each time step $t \in T$. This approach is also known as \textit{teacher forcing}~\cite{williams1989learning}. 

However, using the past targets $y_{t-n}$ and current input $x_t$ to predict $y_{t+1}$ at training time does not emulate the process at test time where the model is instead required to use its own past predictions $\hat{y}_{t-n}$ to generate the sequence. Sequence predictors can suffer from this discrepancy since they are not trained to use past predictions. This leads to a problem where errors compound along a generated sequence at test time, in the worst case leading to errors quadratic in $T$. 
Therefore, learning from high dimensional and sparsely structured patterns at training time to generate sequences from the same distribution at test time is challenging for neural language models (NLMs). Beam search (BS) is often used to address this challenge by greedily searching a subset of the most probable trajectories. However, BS is not suited to the continuous state spaces in NLM, furthermore it requires the use of dynamic programming over large discrete states in the size of the vocabulary. Hence, BS as method for mitigating compounding errors at training time is too slow, not to mention the problem of state transitions being influenced by long-term dependencies (BS depth is typically small). In such cases, learning from a teacher policy with full-supervision throughout training can be sub-optimal. 

We argue that in cases where the learner finds it difficult to learn from the teacher policy alone, sampling alternative inputs that are similar to the original input under a suitable curriculum can lead to improved out-of-sample performance, and can be considered as an exploration strategy. 
Moreover, interpolating this scheme with past predictions further improves performance.
We are motivated by the past findings that using outputs other than that provided by an expert can be beneficial for  generation tasks~\cite{hinton2015distilling,norouzi2016reward}. 

In this paper, we propose a curriculum learning-based method, whereby the teacher policy is augmented by sampling neighboring word vectors within a chosen radius with probability assigned via the normalized cosine similarities between each of the $k$ neighbors and a corresponding input $x_t$. By using pretrained word embeddings as the basis for choosing the neighborhood for each word, we only require to store an embedding matrix $\mat{E} \in \R^{|V|\times d}$ where $|V|$ is the vocabulary size and the $d$ is the dimensionality of the embedding space.
In prediction-based word embeddings, typically $d \ll |V|$, which results in a significantly smaller memory footprint when using $\mat{E}$ than using a transition probability matrix computed from a co-occurrence matrix in $\N^{|V|\times|V|}$.
During training we monotonically increase the replacement probability using a number of simple non-parametric functions. 
We show that performance can be further improved when Nearest Neighbor Replacement Sampling (NNRS) is used in conjunction with scheduled sampling~\cite{bengio2015scheduled}, a standard technique that also addresses the compounding error problem. 

\section{Related Work}
The most relevant prior work to ours is \textit{scheduled sampling}~\cite[SS;][]{bengio2015scheduled}, which also uses an online sampling strategy to reduce compounding errors. To alleviate this problem, SS alternates between $\hat{y}_{t-1}$ and $y_{t-1}$ using a sampling schedule, whereby the probability of using $\hat{y}_{t-1}$ instead of $y_{t-1}$ increases throughout training, allowing the learner to generate multi-step ahead predictions by improving the models' robustness with respect to its own prediction errors. 

\textit{Dataset Aggregation}~\citep[DAgger;][]{ross2011reduction} finds a stationary deterministic policy by allowing the model to first make predictions at test time and then queries the teacher as to what actions it could have taken given the observed errors the model made on the  validation data. DAgger attempts to address compounding errors by initially using an expert policy to generate a set of sequences. 
These samples are added to the dataset $\mathcal{D}$ and a new policy is learned, which subsequently generates more samples appended to $\mathcal{D}$. 
This process is repeated until convergence and the resultant policy $\pi$ that best emulates the expert policy is selected. 
In the initial stages of learning, a modified policy is considered $\pi_i = \beta_i \pi^{*} + (1 - \beta_i)\hat{\pi}_i$, where the expert $\pi^{*}$ is used for a portion of the time (referred to as a mixture expert policy) so that the generated trajectories in the initial stages of training by $\hat{\pi}_i$ at time $i$ do not diverge to irrelevant states. 

Likewise, \textit{Mixed Incremental Cross-Entropy Reinforce}~\cite[MIXER;][]{ranzato2015sequence} applies REINFORCE for text generation.
MIXER also uses a mixed expert policy (originally inspired by DAgger), whereby incremental learning is used with REINFORCE and cross-entropy (CE). 
The main difference between MIXER and SS is that the former relies on past target tokens in prediction and the latter uses REINFORCE to determine if the predictions lead to a sufficiently good return from a task-specific score.
In our proposed method, we instead dynamically interpolate between the model's past predictions, past targets and the additional intermediate step of using past target neighbors and use individual schedules for the past prediction and the past targets' $k$-neighbors.

\section{Methodology}\label{sec:method}

\subsection{Model Description}

In this paper, we consider sequence modelling tasks where the training set contains input-output pairs $(X, Y)$, where $X = x_{1}, x_{2}, \ldots, x_{T}$ is an input sequence of length $T$ and $Y = y_{1}, y_{2}, \ldots, y_{T}$ is its corresponding output sequence of the same length. 
We use the compact notation $x_{1:T}$ to denote the sequence $X = x_{1}, x_{2}, \ldots, x_{T}$.
In language modelling (LM), the task is to predict the next token in the sequence, $x_{t}$, given  $x_{1: t-1}$. For this reason, LM can be seen as a special (unsupervised) instance of the structure prediction problem where $Y = x_{2:T}$ and the input sequence is defined as $X = x_{1:T-1}$.  
Different models have been proposed in the literature for sequence-to-sequence prediction problem such as variants of recurrent neural networks~\cite[RNNs;][]{sundermeyer2012lstm} and Transformers~\cite{vaswani2017attention}. We model these methods collectively as first computing a representation $\vec{h}_{t}$ for the sequence $x_{1:t-1}$ using some (recurrent in the case of RNNs and attention-weighted sum in the case of Transformers) function, $f(x_{1:t-1}; \vec{\theta})$, parameterised by $\vec{\theta}$. Next, the prediction, $\hat{y}_{t}$, of the next target output, $y_{t}$, is modelled as a classification problem where a probability distribution, $p(\cdot)$, over the possible output labels is computed (e.g., using $\mathtt{softmax}$ function) and the label with the maximum probability is selected as $\hat{y}_{t}$. Then, the parameters $\vec{\theta}$ of the generator can be learned such that the log-likelihood of the output sequence $Y$ given by \eqref{eq:ll} is maximized.
\begin{align}\label{eq:ll}
\frac{1}{T}\sum_{t = 1}^{T} \log p(y_t|y_{1:t-1}, X; \vec{\theta})
\end{align}

As a concrete example of this setting, let us consider NLM using an RNN. The recurrent function in the neural network updates its hidden state vector at each time step. Language modelling can be seen as a classification problem where each word in the output vocabulary, $V$, is a candidate label, and we must select the most probable output label (word) as the next prediction. Given the hidden state vector $\vec{h}_{t-1}$, after observing the sequence $x_{1:t-1}$, we can compute the probability of each word $w \in V$ using
$\sigma(\vec{h}_{t}\T\vec{\theta}_{w})$, where $\vec{\theta}_{w}$ is a parameter vector (embedding) corresponding to $w$
and $\sigma$ is the $\mathtt{softmax}$ function.

 Finally, the log-likelihood of the given corpus is computed using \eqref{eq:ll} and stochastic gradient descent (SGD) (or a variant) can be used to find the optimal model parameters for the generator.

Although training can be conducted using the target outputs $Y$, note that they are \emph{not} available during test time.
Therefore, the model uses its current prediction $\hat{y}_{t}$ with hidden state $\vec{h}_t$ to predict $p(\hat{y}_{t+1}|\hat{y}_{t}, h_{t}; \vec{\theta})$. 
As already discussed, this can lead to an accumulation of errors.
The current prediction $\hat{y}_{t}$ can be obtained by either acting greedily and selecting the most probable word or chosen by samping from the distribution $p_{\theta}(y|h_t; \vec{\theta})$, calibrated using the $\mathtt{softmax}$ output. 


\subsection{Addressing Compounding Errors}
Teacher forcing is when a sequence predictor is given full supervision during training, which is the most popular way to train NLMs.
 However, this does not reflect sequence generation at test time since targets are not provided and the model has to rely on its own past predictions to generate the next word. SS addresses this by alternating between $y_{t-1}$ and $\hat{y}_{t-1}$ at training time to encourage the model to improve multi-step ahead prediction used at test time. 
 The trade-off is controlled with probability $\gamma_{i}$ ($\in [0,1]$) at the $i$-th epoch such that with probability $(1 - \gamma_{i})$ the model chooses $\hat{y}_{t-1}$. Herein, we denote the sampling rate as $\epsilon$ for the schedule corresponding to SS and that for NNRS by $\gamma$.
The $\epsilon_i$ coefficients are incrementally decreased during training as a curriculum learning strategy where the model uses more true targets at the beginning of training and gradually changes to using its own predictions during learning. 

There are three ways in which $\epsilon_i$ is set, (1) a linear decay, (2) an exponential decay or (3) an inverse Sigmoid decay.  However, as noted in previous work~\cite{huszar2015not}, SS has the limitation that it can lead to learning an incorrect conditional distribution. Additionally, it is assumed that the teacher is optimal when using full supervision. 
Next, we describe our proposed method that addresses these criticisms while also being sufficiently modular to be used in conjunction with other sampling-based techniques for mitigating exposure bias.

\subsection{Nearest-Neighbor Replacement Sampling}

\paragraph{Defining Neighbors via Embedding Similarity:}

Suppose we have $d$-dimensional pretrained embeddings for words $w \in V$, arranged as rows in an embedding matrix $\mat{E} \in \mathbb{R}^{|V| \times d}$. We denote the embedding of word $w$ by $\vec{w} \in \mathbb{R}^{d}$. For each $w \in V$, we measure its similarity to all other words $w' \in V - \{w\}$ using the cosine similarity, $\cos(w, w')$, given by \eqref{eq:cos}.
\begin{align} \label{eq:cos}
 \cos(w, w') = \frac{\vec{w}\T\vec{w'}}{\norm{\vec{w}}\norm{\vec{w'}}}
\end{align}
We then define the neighborhood similarity matrix, $\mat{N} \in \R^{|V|\times k}$, containing similarities of the top $k$ most similar words $w'$ to $w$ according to $\cos(\vec{w},\vec{w}')$ for all $w \in V$. We denote $\mat{N}_w$ as the row of neighbor embedding similarities corresponding to word $w$. 
Once $\mat{N}$ is computed, we convert each row to a probability distribution in order to sample the neighbors, which we denote as $\tilde{\mat{N}}$ where for a given row corresponding to $w$,  $\sum_{i=1}^{k} \tilde{\mat{N}}_{w, i} = 1$. Hence, a neighbor $w'$ in the $i$-th position of $\tilde{\mat{N}}_{w, i}$ is associated with a probability $p(w'; w, k)$ given by \eqref{eq:nn-prob}, where the sampling probabilities are defined using the $\mathtt{softmax}$ function to  normalize the cosine similarities between pretrained word embeddings.
\par\nobreak
{\small
\begin{align}
	\label{eq:nn-prob}
   	 p(w'| w; k, \tau) = \frac{\exp(\cos(\vec{w},\vec{w}') / \tau)}{\sum_{\vec{u} \in \mat{N}_{w}} \exp(\cos(\vec{u}, \vec{w}) / \tau)}
\end{align}
}%
Here the temperature $\tau$ controls the ``peakiness'' of the distribution, lower $\tau$ corresponding to much higher sampling probability for the closest neighbor and far smaller probabilities for the $k$-th furthest neighbor.
Furthermore, a $w$ that occurs at $y_{t-1}$, can instead be represented as a weighted average (i.e centroid) of its neighbor embeddings $\tilde{\vec{w}}$ as shown in \eqref{eq:mean_neighbor} and subsequently passed as input at $x_t$. 
\par\nobreak
{\small
\vspace{-3mm}
\begin{align}\label{eq:mean_neighbor}
\tilde{\vec{w}} = \frac{1}{k}\sum_{i=1}^{k} p(w'_i|w, k) \vec{w}'_i
\end{align}
}%

In our experiments, $k \approx \log_2(|V|)$ which is sample efficient and speeds up sampling at run time when compared to using the entire vocabulary of size $|V|$.  Note that, in the case where $|V|$ is very large (e.g $|V|>10^4$), efficient k-NN search can be performed using KD-Trees, Metric Trees or Cover Trees algorithms~\cite{kibriya2007empirical}, which has been applied in a similar context~\cite{bollegala2017think}. In the best case, computation can be reduced from $\mathcal{O}(n^2)$ to $\mathcal{O}(n\log n$) at the expense of some accuracy. However, this was not required as $|V| \ll 10^4$ for the datasets we report results on.

During training, samples from top $k$-neighbors are drawn~$\tilde{w} \sim \tilde{\mat{N}}_{w}$, while $\tilde{\mat{N}}_{w}$ can change at the end of each epoch $i \in \Gamma$ based on the validation perplexity $\ell$ by updating $\tau$. This has the effect of shifting the probability mass for the $k$-neighbors sampling probability distribution, proportional to the increase or decrease in $\ell$ from epoch $i$ to $i+1$. 

At the start of training the model is conservative, only sampling the nearest neighbors. As the model begins to learn and minimize the loss on the validation set, it gradually explores more distant $k$ neighbors. If replacing neighbors that have smaller cosine similarity has a similar effect on the validation loss to neighbors with higher cosine similarity, then $\tau$ is increased and the model carries out more exploration. In the opposite case, where validation loss decreases, $\tau$ becomes smaller and the model will only choose the closest of the $k$-neighbors for replacement. Concretely, $\ell_{i}$ is the validation set loss at epoch $i$, and $\ell_{*}$ is the best validation performance up until $i$ number of epochs. When $\ell_{i} < \ell_{*}$, $\tau$ is increased using the update rule shown in \eqref{eq:delta_assign} where $\tau_{i}$ controls the temperature in \eqref{eq:nn-prob} at epoch $i$ and $\varepsilon_{i-1} = (2^{\tau_{i-1}}-1)$. 
\par\nobreak
{\small
\begin{align}\label{eq:delta_assign}
\vspace{-3mm}
       \tau_{i} := 
        \begin{cases}
            \tau_{i-1} + |\tau_{i-1} - \varepsilon_{i-1}|, &  \ell_{i} - \ell_{*} \geq 0,  \\
            \tau_{i}
            - |\tau_{i} - \varepsilon_{i-1}|, & \ell_{i} - \ell_{*} < 0\\
        \end{cases}
\end{align}
}%
Because this procedure depends on $\ell$, the sampling process is directly controlled by the performance metric of interest. 
$\tau \in [0.5, 10]$ is used to prevent  \eqref{eq:nn-prob} becoming too narrow or completely uniform.

This approach can be considered as an auxiliary task optimized for the expected reward.
However, we require a curriculum for NNRS to control the amount of replacement sampling, as $\tau$ would be increased very early on in training because large loss reductions are made where the model is initialized at random. Thus, changes in $\tau$ have a larger effect on the overall sampling process as the chosen schedule monotonically increases the replacement sampling rate over training epochs.

\eqref{eq:sampling} shows the procedure for choosing whether to use past target $y_{t-1}$, past prediction $\hat{y}_{t-1}$ or a sampled neighbor of $y_{t-1}$ and assign it to the next input $x_{t}$.
\par\nobreak
{\small
\vspace{-3mm}
\begin{align}\label{eq:sampling}
    x_t =
\begin{cases}
 \tilde{y}_{t-1} \sim \mathbb{B}(\tilde{\mat{N}}_{y}, \hat{y}), & (\epsilon > \xi_{\text{ss}}) \hspace{0.3em}\& \hspace{0.3em} (\gamma > \xi_{\text{nnrs}}) \\     
\hat{y}_{t-1}, &  (\epsilon > \xi_{\text{ss}}) \hspace{0.3em} \& \hspace{0.3em} (\gamma < \xi_{\text{nnrs}}) \\
 y_{t-1}, & (\epsilon < \xi_{\text{ss}}) \hspace{0.3em} \& \hspace{0.3em} (\gamma > \xi_{\text{nnrs}}) \\          
    \end{cases}
\end{align}
}%

We start by sampling uniformly at random $\xi_{\text{ss}} \sim \mathrm{Uniform}(0, 1)$ and $\xi_{\text{nnrs}} \sim \mathrm{Uniform}(0, 1)$ for each sample $Y$ within a training mini-batch $B^i$ in $\mathbf{B}_{tr} = \{B^1, B^2, B^i, \dots B^K\}$. Here, $\xi_{\text{ss}}$ is used as a threshold to decide which $y \in Y$ are used for SS and $\xi_{\text{nnrs}}$ sets a threshold for NNRS when both are used in conjunction. When both $\epsilon$ and $\gamma$ choose the same tokens for sampling, we choose one at random with equal probability. This point becomes more relevant towards the end of training when the schedule outputs a high sampling rate.
For faster inference, we fix the sampled indices corresponding to $y \in Y$ for all samples in $Y$. This means NNRS can be performed in parallel on all mini-batch samples in $\mathbf{B}_{tr}$. Below, we have described the use of NNRS in conjunction with another popular sample-based method to demonstrate its modularity. However, NNRS can also be used standalone as we see later in our experiments. 
In fact, NNRS can be expanded upon to learn which neighbors to sample as we describe next.

\paragraph{Gumbel Softmax Neighbor Sampling:}
An alternative to updating the temperature via the incremental update rule in \eqref{eq:sampling} is to use a straight-through estimator that allows us to differentiate through drawn samples $p(w'|w; k, \tau)$. 
This can be achieved using the Gumbel-Softmax~\cite[GS;][]{jang2016categorical} trick. 
For NNRS, we refer to this as Gumbel-Softmax Neighbor Sampling (GSNS). The GS allows us to sample and backpropogate through $\tilde{\mat{N}}$ by reparameterizing the sampling process as $\mathbf{d}\tilde{\mat{N}}_w/\mathbf{d}\alpha$ where $\alpha$ is a multinomial distribution.
\eqref{eq:concrete_dist_tree} shows the GS where each componentwise Gumbel noise $\kappa \in [1..,k]$ added to the original distribution  $p(\tilde{\mat{N}}_{w})$ for $w$, we find $\kappa$ that maximizes $G_w^{\kappa}:= \log \alpha_w^{\kappa} - \log(-\log U_w^{\kappa})$ and then set $D_w^{\kappa}=1$ and the remaining elements $D_w^{\neg \kappa} = 0$ (i.e a one-hot vector).  Here, $U_w^{\kappa} \sim \text{Uniform}(0, 1)$ and $\alpha_w^{\kappa}$ is drawn from the discrete distribution $D \sim \text{Discrete}_{w}(\alpha)$. 
We then sample $p(w'|w; k, \tau)$ as in \eqref{eq:concrete_dist_tree} after updating $\nabla_{p(\mat{N}_w)}\ell_{i}$. 
\par\nobreak
{\small
\begin{align}\label{eq:concrete_dist_tree}
p(w'|w; k, \tau) = \frac{\exp((\log \alpha_w^{\kappa} + G_w^{\kappa})/\tau)}{\sum_{i=1}^{k} \exp((\log \alpha_{w}^i + G_{w}^i)/\tau)}
\end{align}
}%

This allows us to draw samples from a Gumbel distribution while performing gradient updates on $p(\mat{N}_{w})$.
 In contrast to NNRS, which performs updates according to the validation perplexity at the end of the of each training epoch, we instead update the neighbor distribution throughout training. 
 However, similar to before, the curriculum still controls the amount of GSNS updates that are performed. 
 Therefore the gradient updates are only for the tokens, which are chosen for GSNS according to the curriculum schedule. 
We test $\tau = 0.5$ as a constant during training, 
We set an upper bound of $\tau = 10$, which corresponds to a heavy dampening of \eqref{eq:concrete_dist_tree}.
Small $\tau$ early avoids high variance in gradient updates, even-though we would expect the curriculum to use NNRS less early on.

We expect the output to be less sensitive to perturbations in the input because the input is locally bounded by the space occupied by the $k$ nearest neighbors of the target $\tilde{y}_{t}$. Likewise, $\tilde{y}_{t}$ can be considered as emulating the problem of compounding errors, since the conditional probability $p(y_{t+1}|x_{1:t},\tilde{y}_{t};\theta)$ is conditioned on the sampled neighbor $\tilde{y}_{t}$ instead of the true target $y_{t}$.
In cases where the model finds it difficult to transition from using $y_{t-1} \to \hat{y}_{t-1}$, interpolating with the neighborhood samples $\tilde{y}$ can provide a smoother policy ($y \to \tilde{y} \to \hat{y}$). This smoothing assigns some mass to unseen transitions (similar to Laplacian smoothing), bounded by $k$ neighbors, which is directly proportional to transition probabilities. 
We now consider curriculum schedules to monotonically increase both $\gamma$ and $\gamma_{\mathrm{nnrs}}$, corresponding to the sampling rates for SS and NNRS respectively and aim to identify schedules that help mitigate compounding errors by controlling the amount of  exploration of neighbors throughout training.

\textbf{\textit{Objective Definition}}
We note that, in SS, the objective is not a strictly proper scoring rule (i.e the maximum of the function is unique) since the objective is dependent on the models own distribution. However, in NNRS, we do not use the model's distribution but instead neighbors of past targets which are not subject to change throughout training. In other words, the sampling procedure is chosen by a predefined curriculum and therefore is independent of the model, thus the maximum of this scoring function remains unique.  




In the case of a static sampling rate for $\epsilon$ and $\gamma$, we can define the log-likelihood loss with SS and NNRS in terms of KL-divergences, shown in \eqref{eq:ss_nnrs}. 
$P_{x_{t}}$ and $Q_{x_{t}}$ are the marginal distributions for input token $x_{t}$, while $P_{x_{t}|x_{t-1}=h}$ and $Q_{x_{t}|x_{t-1}=h}$ are the conditional distributions.  SGD with cosine annealing~\cite{reddi2018convergence} of the learning rate is also considered for optimizing the model. 
\par\nobreak
{\small
\vspace{-3mm}
\begin{align}\label{eq:ss_nnrs}
\begin{gathered}
	D_{\text{SS-NNRS}}[P||Q] = \text{KL}[P_{x_{t-1}}||Q_{x_{t-1}}] +\\
     \underbrace{(1 - \epsilon)\mathbb{E}_{h \sim Q_{x_{t-1}}}\text{KL}[P_{x_{t-1}}||Q_{x_{t}|x_{t-1}=h}]}_\text{SS} + \\ \epsilon \underbrace{\mathbb{E}_{h \sim P_{x_{t-1}}}\text{KL}[P_{x_{t}|x_{t-1}}|| Q_{x_{t}|x_{t-1}}] }_\text{SS} + \\
    \underbrace{(1 - \gamma) \mathbb{E}_{h \sim Q_{x_{t-1}}} \text{KL}[P_{x_{t+1}}||Q_{x_{t}|x_{t-1}=h}]}_\text{k-NN replacement sampling}+\\ 
 \underbrace{\gamma \mathbb{E}_{h \sim P_{x_{t-1}}} \text{KL}[P_{x_{t}|x_{t-1}}|| Q_{x_{t}|x_{t-1}}]}_\text{k-NN replacement sampling}
 \end{gathered}
\end{align}
}%

In practice, we only need to use the CE loss when using NNRS. The aforementioned KL analysis suggests that when using the CE loss, we the scoring function is proper, unlike the SS objective. At this point, we have defined the whole process to carry out NNRS and summarize the training with NNRS in \autoref{alg:knn_alg}.


\begin{algorithm}[!t]
\small
\KwIn{Training epochs $\Gamma$. Sentence length $T$ for training mini-batch training data $\mathbf{B}_{\text{tr}} = \{B_{\text{tr}}^1, B_{\text{tr}}^2, \dots, B_{\text{tr}}^K\}$ and validation data $\mathbf{B}_{\text{val}} = \{B_{\text{val}}^1, B_{\text{val}}^2, \dots, B_{\text{val}}^R\}$. An RNN $f_\theta(\cdot, \cdot)$ parameterized by $\theta$. Curriculum schedule functions $g(\cdot)_{ss}, g(\cdot)_{nnrs}$.}

Initialize the RNN parameters $\theta$ at random \\
Define k-NN embedding similarity for vocabulary $V$ using \eqref{eq:cos} and normalize to a probability distribution $p(\tilde{\mat{N}})$ using \eqref{eq:nn-prob} \\
Set $\tau=0.1, i=0, \xi=0$,  $\ell_{*}=|V|$  \\

\ForEach {$i \in \Gamma$}
{
Set $\epsilon = g_\text{ss}(i), \gamma_{\text{nnrs}} = g_\text{nnrs}(i), \ell_{i}=0 $ \\

\ForEach {$B^k \in \mathbf{B}_{tr}$}
{
    \For {step $t = 1 \to T_{k}$}
    	{
    	    Sample $\xi_{\text{ss}} \sim \mathbb{U}(0, 1)$ and $\xi_{\text{nnrs}} \sim \mathbb{U}(0, 1)$ \\
    		Sample $\vec{x}_{t}$ using \eqref{eq:sampling} and input\\
    		    \quad $\hat{\vec{y}}_t, h_{t} = f_{\theta}(\vec{x}_t, \vec{h}_{t-1})$ \\
    	}
    	Compute cross-entropy loss \\
        \quad $\ell_{\text{tr}} = \mathcal{L}_{\text{ce}}(\hat{\mat{Y}}, \mat{Y})$ \\
    	Update $\theta$ \\
    	    \quad $\theta \from \eta \theta + (1 - \eta) \nabla_{\theta}\log \ell_{\text{tr}} $ \\
    }
	Compute $\ell_{i}$ by repeating above on $\mathbf{B}_{\text{val}}$ without updating $\theta$ \\
	Update $\ell_{*}$ if $\ell_{i} < \ell_{*}$ \\
	\quad \textbf{if} using the Gumbel-Softmax \\
	    \quad\quad Update $\mat{N} := \beta \mat{N} - (1 - \beta) \nabla_{\tilde{\mat{N}}} \ell_{*} $ \\
	\quad \textbf{else} \\
		\quad\quad Update $\mat{N}$ by changing $\tau$ based on the difference \\ 
		\quad\quad between $\ell_{*}$ and $\ell_{i}$ using  \eqref{eq:delta_assign}\\
	\quad Renormalize $p(\mat{N})$ using the $\mathtt{softmax}$
	
}

\caption{NNRS-based Training}
\label{alg:knn_alg}
\end{algorithm}

\section{Experiments}

\begin{table*}
\centering
\def\arraystretch{1.0}
\begin{small}
\resizebox{1.0\linewidth}{!}{%

\begin{tabular}{c|cccc|cc|cc|cc|cc?cc|cc|cc|cc}
 \toprule[2.pt]

Configuration  & \multicolumn{4}{c}{Parameter Setting} & \multicolumn{8}{c?}{Wiki-102} & \multicolumn{8}{c}{Penn-Treebank}\\

\multicolumn{5}{c|}{} & \multicolumn{2}{c|}{Linear} & \multicolumn{2}{c}{S-Shaped Curve} & \multicolumn{2}{|c}{Exponential Increase}& \multicolumn{2}{|c?}{Static}  & \multicolumn{2}{c|}{Linear} & \multicolumn{2}{c}{S-Shaped Curve} & \multicolumn{2}{|c}{Exponential Increase}& \multicolumn{2}{|c}{Static}\\

\midrule

& $\epsilon_s$ & $\epsilon_e$ & $\gamma_s$ & $\gamma_e$ & Valid & Test & Valid & Test & Valid & Test & Valid & Test & Valid & Test & Valid & Test & Valid & Test & Valid & Test\\
\midrule 


No Sampling & - & - & - & - & 140.28 & 128.78 & 140.28 & 128.78 & 140.28 & 128.78 & 140.28 & 128.78 & 76.25 & 71.81 & 76.25 & 71.81 & 76.25 & 71.81 & 76.25 & 71.81 \\
\midrule
TPRS-1 & 0 & 0 & 0 & 0.2 & \textbf{\emph{143.78}} & \textbf{\emph{131.18}} & \textbf{\emph{137.69}} & \textbf{\emph{127.05}} & 137.63 & 136.88 & 136.31 & 126.49 & \textbf{\emph{81.42}} & \textbf{\emph{76.58}} & \textbf{\emph{83.40}} & \textbf{\emph{81.22}} & 77.56 & 73.02 & \textbf{\emph{81.45}} & \textbf{\emph{80.36}}\\
TPRS-2 & 0 & 0 & 0 & 0.3 & 154.93 & 144.02 & 146.92 & 137.07 & \textbf{\emph{137.11}} & \textbf{\emph{125.41}} & \textbf{\emph{137.10}} & \textbf{\emph{125.95}} & 96.91 & 94.30 & 88.79 & 86.17 & \textbf{\emph{77.63}} & \textbf{\emph{72.80}} & 91.73 & 84.48\\
TPRS-3 & 0 & 0 & 0 & 0.5 & 159.11 & 148.41 & 148.10 & 139.58 & 138.46 & 127.97 & 138.53 & 129.87 & 97.62 & 94.78 & 88.46 & 86.05 & 76.60 & 72.77 & 98.31 & 95.23\\


\midrule

NNRS-1 & 0 & 0 & 0 & 0.2 & \textbf{\emph{142.53}} & \textbf{\emph{130.45}} & \textbf{\emph{137.08}} & \textbf{\emph{126.82}} & 136.81 & 136.11 & 135.06 & 136.02 & \textbf{\emph{80.91}} & \textbf{\emph{76.70}} & \textbf{\emph{83.17}} & \textbf{\emph{79.62}} & \textbf{\emph{76.00}} & \textbf{\emph{72.19}} & \textbf{\emph{82.23}} & \textbf{\emph{80.03}}\\
NNRS-2 & 0 & 0 & 0 & 0.3 & 154.50 & 143.13 & 149.69 & 136.98 & \textbf{\emph{136.83}} & \textbf{\emph{125.53}} & 136.34 & 125.84 & 96.66 & 93.18 & 88.68 & 85.29 & 77.12 & 73.05 & 92.12 & 84.29\\
NNRS-3 & 0 & 0 & 0 & 0.5 & 158.30 & 147.13 & 148.02 & 137.34 & 137.56 & 127.61 &  \textbf{\emph{132.62}} &  \textbf{\emph{121.50}} & 96.95 & 93.15 & 88.21 & 84.47 & 75.91 & 72.46 & 97.55 & 94.79\\

\midrule
SS-1 & 0 & 0.2 & 0 & 0 & 139.31 & 128.50 & 143.82 & 131.08 & 137.72 & 126.46 & 135.55 & 122.61 & 83.63 & 80.03 & \textbf{\emph{82.33}} & \textbf{\emph{79.02}} & 76.71 & 73.32 & 74.50 & 70.20\\
SS-2 & 0 & 0.3 & 0 & 0 & 137.78 & 126.00 & 138.39 & 126.80 & \textbf{\emph{135.89}} & \textbf{\emph{125.03}} & 131.29 & 121.52 & \textbf{\emph{94.74}} & \textbf{\emph{79.82}} & 84.42 & 80.03 & 76.43 & 73.46 & \textbf{\emph{74.56}} & \textbf{\emph{70.25}}\\
SS-3 & 0 & 0.5 & 0 & 0 & \textbf{\emph{135.14}} & \textbf{\emph{124.29}} & \textbf{\emph{136.88}} & \textbf{\emph{125.41}} & 136.98 & 125.96 & \textbf{\emph{131.28}} & \textbf{\emph{121.51}}  & 92.48 & 88.92 & 85.87 & 82.37 & 76.41 & 73.15 & 74.11 & 69.48\\
SS-4 & 0 & 0.8 & 0 & 0 & 140.97 & 130.00 & 141.13 & 129.99 & 138.09 & 127.23 & 134.32 & 122.86 & 92.94 & 88.85 & 87.29 & 84.27 & \textbf{\emph{76.22}} & \textbf{\emph{72.98}} & 74.02 & 70.23\\

\midrule
SS-NNRS-1 & 0 & 0.2 & 0 & 0.2 & 139.32 & 128.50 &  \textbf{\emph{141.57}} & \textbf{\emph{129.67}} & 137.73 & 126.47 & 135.55 & 122.60 & \textbf{\emph{83.62}} & \textbf{\emph{80.03}} & \textbf{\emph{81.99}} & \textbf{\emph{77.86}} & 76.71 & 73.31 & 74.50 & 70.20\\
SS-NNRS-2 & 0 & 0.3 & 0 & 0.3 & 137.78 & 126.01 & 147.34 &135.61 & 136.02 & 125.73  & 137.73 & 126.47 & 95.39 & 92.18  & 87.97 & 84.64 & 75.43 & 72.46 & 74.64 & 70.43\\

SS-NNRS-3 & 0 & 0.5 & 0 & 0.2 & \textbf{\emph{134.79}} & \textbf{\emph{123.90}} & 149.00 & 137.97 & \textbf{\emph{135.82}} & \textbf{\emph{124.72}} &  \cellcolor{black!20}{\emph{130.95}} & \cellcolor{black!20}{\emph{120.76}} & 95.90 & 92.96 & 88.56 & 84.96 & \textbf{\emph{74.83}} & \textbf{\emph{70.95}} & \cellcolor{black!20}\emph{72.89} & \cellcolor{black!20}69.06\\

SS-NNRS-4 & 0.2 & 0.5 & 0.2 & 0.5 & 150.97 & 138.59 & 146.01 & 134.67 &  135.88 & 126.02 & 123.84 & 121.98 & 96.78 & 93.20 & 88.56 & 87.06 & 76.41 & 73.15 & 74.42 & 69.88\\

SS-NNRS-5 & 0 & 0.5 & 0 & 0.5 & 136.22 & 125.70 & 151.39 & 138.44 & 137.02 & 125.84 & 132.17 & 121.86 & 97.12 & 93.90 & 90.37 & 86.25 & 76.02 & 72.31 & 74.08 & 70.79\\

SS-NNRS-6 & 0 & 0.8 & 0 & 0.2 & 148.96 & 130.01 & 141.14 & 129.99 & 138.09 & 127.24 & 134.32 & 122.89 & 95.91 & 92.87 & 81.89 & 78.54 & 74.74 & 70.64& 74.02 & 70.23 \\

SS-NNRS-7 & 0.2 & 0.8 & 0.2 & 0.5 & 155.17 & 148.21 & 149.58 & 137.83 & 135.45 & 124.82 & 131.09 & 121.43 & 96.77 & 93.64 & 88.27 & 85.04 & 76.35 & 72.74 & 74.43 & 70.12\\

\bottomrule[2.pt]
\end{tabular}%
}
\end{small}
\captionsetup{justification=centering}
   \caption{Test Perplexity for a 2-hidden layer LSTM~\shortcite{hochreiter1997long} using Transition Probability Sampling (TPRS), SS and Nearest-Neighbor Replacement Sampling (NNRS) with linear, s-shaped curve and exponential sampling functions}
  \label{tab:nsr_results}
\end{table*}



\begin{figure}
\centering
\captionsetup{justification=centering}
 \includegraphics[scale=0.5]{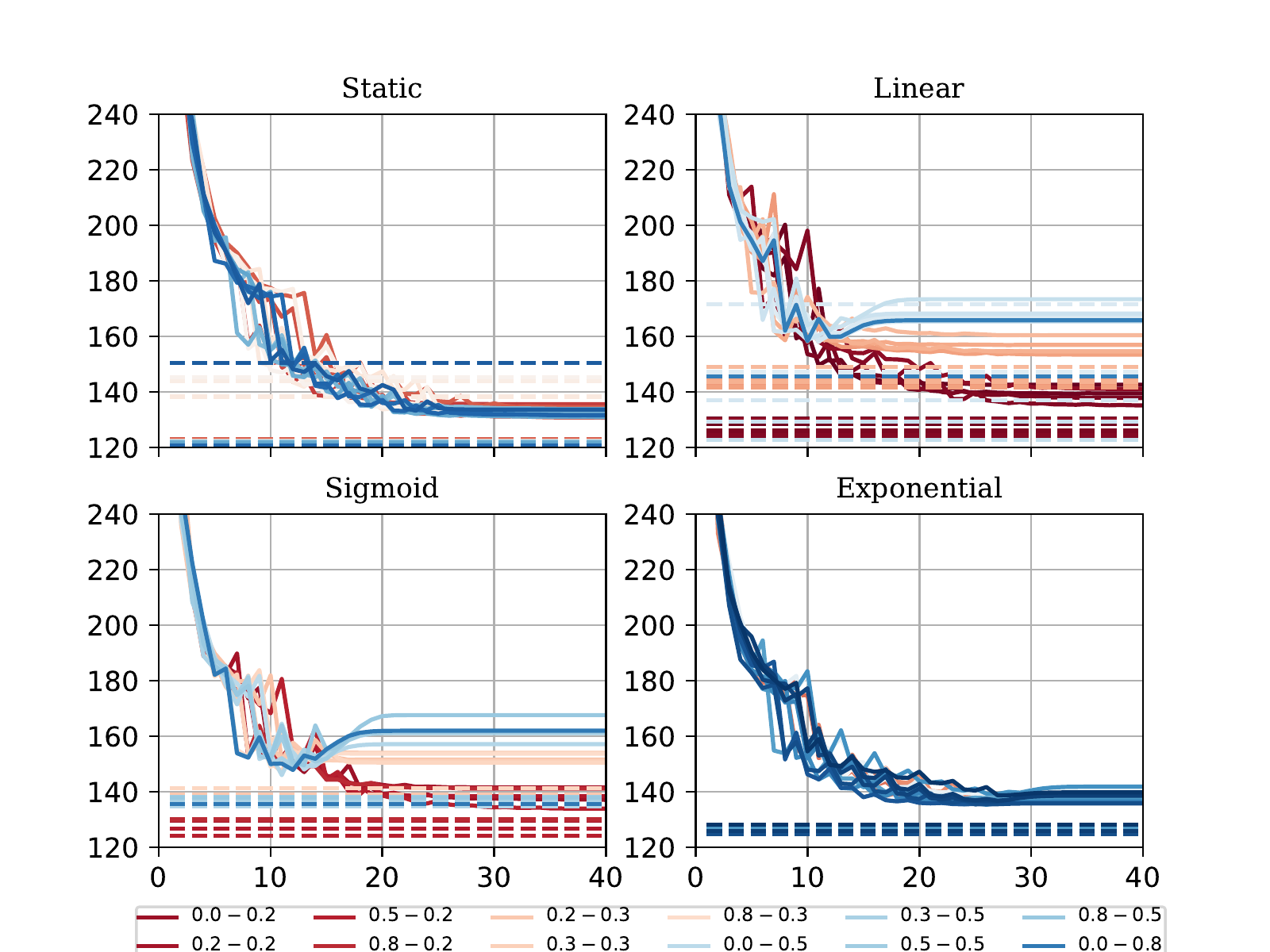}
 \caption{Perplexity on WikiText2: SS and NNRS (best viewed in color)}\label{fig:wiki2_lc_ss}
\end{figure}

We use the Penn-Treebank~\citep[(PTB)][]{marcus1993building} and WikiText-2~\cite{bradbury2016quasi} LM datasets for our expeiments and evaluate the performance of each model using perplexity the defacto standard (please see supplementary for dataset and training details).

\subsection{Experimental Results}


\autoref{fig:wiki2_lc_ss} shows the perplexity scores on WikiText-2 for the different parameter settings for $\epsilon$ and $\gamma$ over 40 training epochs. Solid lines indicate validation perplexity scores throughout training and dashed horizontal lines indicate corresponding test perplexity score. We find that most learning is carried out after 20 epochs given that the learning rate is initially high ($\alpha=20$) and annealed using cosine annealing. 
Based on the validation set performance, we found the optimal settings to be $\epsilon = 0.5$ and $\gamma = 0.2$ with a static probability sampling rate.
Additionally, using an exponential sampling rate outperforms the linear and sigmoid functions. It appears that this is because the exponential function carries out the majority of sampling late in the training when the model has reached learning capacity from the teacher policy.

For PTB, static and exponential settings for both SS and NNRS show best performance and converge quicker. In all cases, convergence for all functions behaves as expected for $\{\sigma, \gamma\} < 0.3$. We also see the same trend as with WikiText-2, where the exponential function allows for higher sampling rates when compared to linear and sigmoid functions. Again, this suggests that NNRS is most effective near convergence, as the sampling probability exponentially increases while the validation perplexity begin to plateau over epochs.  

\paragraph{Schedule Parameter Grid Search Results}
\autoref{tab:nsr_results} shows the results of the model with varying $\epsilon$ and $\gamma$ upper and lower thresholds using a linear, s-curve, exponential and static sampling strategy for all tested datasets. Here, subscript $s$ for $\epsilon_s, \gamma_s$ are the initial probabilities at $i=0$ and subscript $e$ for $\epsilon_e, \gamma_e$ denote end probabilities $i=\Gamma$. Bolded results are those which perform within the cell that are contained (aggregated by configuration and type of schedule e.g SS-1 - SS4 for linear) and shaded cells corresponds to the best performing configuration and schedule for either Wikitext-2 or PTB. These experiments describe results using the simple update rule described in \eqref{eq:delta_assign}.

\begin{table*}[ht]
	\centering
	\resizebox{1.\textwidth}{!}{%
	\begin{tabular}{lc|cc|cc|cc|cc|cc?cc|cc|cc|cc|cc}
		\toprule[2.pt]
		\textbf{Eval.}  & \textbf{Model} &   \multicolumn{2}{c}{\textbf{MLE}} & \multicolumn{2}{c}{\textbf{TPRS}} & 	\multicolumn{2}{c}{\textbf{NNRS}} & \multicolumn{2}{c}{\textbf{SS}} & \multicolumn{2}{c?}{\textbf{SS-NNRS}} &    \multicolumn{2}{c}{\textbf{MLE}} & \multicolumn{2}{c}{\textbf{TPRS}} & 	\multicolumn{2}{c}{\textbf{NNRS}} & \multicolumn{2}{c}{\textbf{SS}} & \multicolumn{2}{c}{\textbf{SS-NNRS}} \\
        \midrule
        \parbox[t]{2mm}{\multirow{4}{*}{\rotatebox[origin=c]{90}{\textbf{BLEU-4}}}}\\
		& LSTM & 7.84 & 8.75 & 7.14 & 6.88 & 7.38 & 7.83 & 10.06 & 9.61 & 7.72 & 7.88 & 6.11 & 6.62 & 5.74 & 5.31 & \textbf{\emph{4.49}} & \textbf{\emph{4.31}} & 6.02 & 5.66 & 5.03 & 4.79 \\
		& GRU & 8.22 & 9.53 & 7.33 & 7.41 & \textbf{\emph{7.88}} & \textbf{\emph{7.23}} & 9.78 & 9.90 & 7.27 & 7.50 & 6.03 & 6.31 & 5.45 & 5.03 & 4.79 & 4.67 & 5.73 & 5.50 & 5.26 & 4.61 \\
		& Highway & 9.13 & 9.26 & 8.04 & 7.93 & 8.29 & 9.14 & 11.11 & 10.51 & 8.73 & 8.18 & 6.84 & 7.04 & 6.34 & 6.13 & 5.77 & 5.59 & 5.63 & 5.10 & 6.08 & 5.28 \\
		\midrule
	    \parbox[t]{2mm}{\multirow{4}{*}{\rotatebox[origin=c]{90}{\textbf{WMD}}}}\\
		
		& LSTM & 0.41 & 0.40 & 0.48 & 0.41 & 0.35 & 0.30 & 0.34 & 0.32 & \emph{\textbf{0.29}} & \emph{\textbf{0.28}} & 0.51 & 0.46 & 0.44 & 0.36 & 0.41 & 0.36 & 0.43 & 0.42 & \emph{\textbf{0.36}} & \textbf{\emph{0.34}}\\

        & GRU & 0.43 & 0.42 & 0.47 & 0.36 & 0.38 & 0.34 & 0.39 & 0.33 & 0.36 & 0.31 & 0.53 & 0.49 & 0.43 & 0.34 & 0.42 & 0.38 & 0.40 & 0.39 & \emph{0.38} & \emph{0.35}\\
        
        & Highway & 0.48 & 0.45 & 0.51 & 0.53 & 0.52 & 0.54 & 0.59 & 0.31 & 0.36 & 0.36 & 0.53 & 0.51 & 0.48 & 0.40 & 0.39 & 0.39 & 0.42 & 0.40 & \emph{0.40} & \emph{0.38}\\
		
		\bottomrule[2.pt]
	\end{tabular}%
	}
	\caption{PTB (left) and WikiText-2 Self-BLEU4 and Self-WMD Validation/Test  scores}
	\label{tab:diversity}
\end{table*}
NNRS and SS used in conjunction yield the best performance in most cases for both datasets. Best performance for both Wikitext-2 and PTB is found with $\epsilon \in [0,0.5]$ and $\gamma \in [0,0.2]$, and slightly improves over only using SS.  For PTB, $\gamma_e = 0.2$ performs the best for linear and sigmoid functions, $\gamma = 0.5$ for exponential and static sampling rates and overall a constant sampling rate. 
Although, the exponential schedule is better than the sigmoid and linear scheduled for both $\epsilon$ and $\gamma$ on both datasets. It seems that this is because the majority of neighbor replacements are towards the end of the training when the learner has already reached the full capacity in what can be learned from the teacher policy.
This coincides with the theoretical guarantees of using an exponential decay schedule provided in DAgger~\cite{ross2010efficient}.

Overall, interpolating between NNRS and SS produces the best performance on average for both datasets. A low constant sampling rate yields best performance for both datasets. Controlling the temperature $\tau$ based on the validation perplexity allows all $k$-neighbors to be explored. When compared to using no sampling, there is an 8 point perplexity decrease using our approach on WikiText-2 and a 2.75 test perplexity decrease on PTB. In comparison to Transition Probability Replacement Sampling (TPRS), NNRS slightly improves while requiring less memory to store embeddings. 

\subsection{Evaluation}
Apart from improving over baselines in terms of perplexity, we identify what tradeoff is incurred between text generation quality and diversity for our proposed method. 
In these experiments we use the best performing $\epsilon$ and $\gamma$ settings found post-analysis for MLE ($\gamma_s=\gamma_e=0$ and $\epsilon_s=\epsilon_e=0$), SS ($\epsilon_s=0, \epsilon_e=0.5$), TPRS, NNRS ($\gamma_s=0, \gamma_e=0.2$ with a static sampling rate, for SS $\epsilon_s=0, \epsilon_e=0.5$) and SS-NNRS (combine previous $\gamma$ and $\epsilon$ settings) trained models. 

\autoref{tab:diversity} shows the self-BLEU~\cite{zhu2018texygen} scores, where higher scores correspond to less diversity in the predictions. Similarly, we also define a self-WMD to measure the semantic diversity which is computed by averaging the average Word Movers Distance~\cite[WMD;][]{kusner2015word} between $\ell_2$-normalized embeddings (the same GoogleNews pretrained $\mathtt{skipgram}$ embeddings~\cite{mikolov2013distributed} that we used to define neighbors) of predicted words and target words in the test set. The WMD scores are in [-1, 1] as the cosine similarity is used as the WMD measure. After computing WMD, the scores are normalized to be in the range of $[0, 1]$. Hence, low self-WMD similarity scores closer to 0 correspond to high diversity. In contrast, when both self-BLEU and self-WMD are high in \autoref{tab:quality}, this signifies high quality. We note that we only consider terms in the validation and test sets that also occur in the training set as including them in the evaluation deflates quality scores.  


\paragraph{Text Generation Diversity}
\newcite{zhu2018texygen} measured the diversity of text generation by computing BLEU scores between a collection of generated samples (referred to as self-BLEU). Self-BLEU can help identify mode collapse in generative models (e.g generative adversarial networks~\cite{goodfellow2014generative}). In the context of uncondtional and conditional NLM, this can correspond to only predicting a subset of the vocabulary whereby the model fails to predict other \textit{modes} (i.e words) which may occur due to the infrequent use of some terms in the long tail of the frequency distribution.  
In \autoref{tab:diversity}, we calculate Self-BLEU4 on both the validation and test sets for the best performing models on PTB and WikiText-2. 
To speed up computation, we avoid computing the self-BLEU4 between all generated samples but instead within each mini-batch and then average over all mini-batches. 
We find that self-WMD in particular shows an increase in diversity for NNRS when compared to standard maximum likelihood training. Although coherently evaluating diversity is difficult, it at least suggests that NNRS is generating sentences that are semantically similar when using the aforementioned GoogleNews pretrained $\mathtt{skipgram}$ embeddings. 


\begin{table}
	\centering
	\resizebox{.47\textwidth}{!}{%
	\begin{tabular}{lc|cc|cc|cc|cc|cc}
		\toprule[2.pt]

		\textbf{}  & \textbf{Model} &   \multicolumn{2}{c}{\textbf{MLE}} & \multicolumn{2}{c}{\textbf{TPRS}} & 	\multicolumn{2}{c}{\textbf{NNRS}} & \multicolumn{2}{c}{\textbf{SS}} & \multicolumn{2}{c}{\textbf{SS-NNRS}} \\
        \midrule
        \parbox[t]{2mm}{\multirow{4}{*}{\rotatebox[origin=c]{90}{\textbf{BLEU4}}}}\\
		& LSTM & 7.87 & 8.28 & 9.24 & 8.16 & 11.81 & 11.26 & 10.93 & 10.53 & \emph{\textbf{11.62}} & \emph{\textbf{11.20}} \\
		& GRU & 9.39 & 8.58 & 9.49 & 10.67 & 11.35 & 11.04 & 10.98 & 11.60 & 12.14 & 11.79 \\
		& Highway & 9.03 & 8.56 & 9.72 & 9.40 & 10.81 & 10.37 & 11.24 & 11.96 & \emph{\textbf{13.75}} & \emph{\textbf{14.03}} \\
		\midrule
		
		 \parbox[t]{2mm}{\multirow{4}{*}{\rotatebox[origin=c]{90}{\textbf{WMD}}}}\\
		& LSTM & 0.72 & 0.84 & 0.85 & 0.93 & 0.91 & 0.88 & 0.89 & 0.91 & \emph{\textbf{0.95}} & \emph{\textbf{0.93}}\\
		& GRU & 0.72 & 0.72 & 0.67 & 0.63 & 0.70 & 0.69 & 0.70 & 0.69 & \emph{0.82} & \emph{0.84}\\
		& Highway & 0.70 & 0.69 & 0.74 & 0.72 & 0.78 & 0.76 & 0.72 & 0.75 & \emph{0.79} & \emph{0.80}\\
		\bottomrule[2.pt]
	\end{tabular}%
	}
	\caption{WikiText-2 BLEU-4 \& WMD Quality Scores}
	\label{tab:quality}
\end{table}

\paragraph{Text Generation Quality}
Following prior work by~\newcite{yu2017seqgan}, we compute BLEU between the predictions and the corresponding targets as shown in \autoref{tab:quality}. 
Similar to the the diversity evaluation, WMD is computed between the pretrained embeddings of the predicted text and target text. 
We find that SS-NNRS leads to slightly more diverse sentences given BLEU-4, but when taking sentence-level semantic similarity using WMD, we find that the improvement is more evident as WMD takes into account the similarity between $k$ local neighbors and targets in the NNRS method, unlike BLEU-4 which treats all incorrect predictions as equally bad. For larger $k$, diversity is increased in expense for generation quality, while $k$=0 corresponds to ML training.

\subsection{Discussion}
In closing, we summarize our findings based on the above LM experiments on PTB and WikiText-2.

Firstly, SS-NNRS with $\epsilon=0 \to 0.5$ and $\gamma= 0 \to 0.2$ has shown the best performance on both PTB and WikiText-2. From this, we conclude that maximum likelihood training is more suitable early on training to allow the model to be robustness enough before learning from synonymous neighboring tokens. 
Secondly, we find that NNRS improves over SS when used in isolation, most notably using a linear schedule. We posit that using past predictions in the intermediate stages of training can be too difficult for the model to adapt to since the model predictions early on are poor and hence recorrecting EB is more difficult. Learning from the targets neighbors is an easier alternative before beginning to increase the sampling rate for choosing past predictions. 

Thirdly, sampling neighbors using the whole transition probability matrix leads to poor results even with relatively low sampling rates. 
Therefore, not only are lower-dimensional embeddings more efficient in sampling during training, but also leads to better performance in the practical setting with a set number of epochs. This related to the tradeoff between exploration and exploitation, where TPRS exploration is too much and therefore leads to a decrease in performance, as reflected in \autoref{tab:nsr_results}.

Lastly, larger test perplexity reductions are found for WikiText-2 when using NNRS. Unlike PTB which has truncated vocabulary size of 10,000, WikiText-2 does not have a reduced vocabulary set and therefore rare terms are kept in the vocabulary, which contains 50,000 words. Hence, NNRS is particularly useful for improving the predictions of less frequent words, particularly when the dataset is relatively small which leads to poor approximations of the transition probability matrix that is necessary when using TPRS. 


\section{Conclusion}

We presented a sampling strategy for curriculum learning to mitigate compounding errors in neural sequence models. Consistent performance improvements are made over standard maximum likelihood training, particularly for sampling functions that generate monotonically increasing sampling rates that are inversely proportional to the slope of the validation performance. This is empirically demonstrated when comparing a standard 2-hidden layer LSTM (with identical hyperparameter settings) with (1) no sampling strategies, (2) a baseline transition probability replacement sampling, (3) the proposed NNRS technique, (4) scheduled sampling and (5) a combination of (3) and (4).

Overall, for both datasets we find that an exponential and static sampling probability outperform sigmoid and linear schedules. Concretely, a schedule that has a high sampling rate too early in training leads to a performance degradation whereas sampling with high probability towards the end of training can improve generalization. Lastly, the test set perplexity scores increase when using NNRS.

\bibliographystyle{acl_natbib}
\bibliography{acl2020}

\end{document}


\maketitle

\section{Structured Output Prediction Methods}
In the past decade there has been a significant effort to improve models for structured output prediction using methods from supervised learning and reinforcement learning. This section expands on related work to supplement to existing cited works in the paper.  

\subsection{Imitation Learning Approaches}
\textit{SEARN}~\cite{daume2009search} treats structured prediction as a search problem by decomposing structured prediction problems into a set of binary classifications without the need to decompose the loss function or the feature functions. A local classifier is used to sequentially (or independently) predict tokens (no actual search algorithm is involved) using past inputs and past predictions. At training time, a sequence of decisions is made to perform search given an expert policy. The new policy is generated from the examples provided by this policy, in which the old policy is interpolated with the new one.

\newcite{he2012imitation} propose \textit{Coaching} to overcome the difficulty of finding an optimal policy $\pi^{*}$ when there is a significant difference between the solution space and the experts ability, which makes it difficult to produce low error on the training data. A coach is used to learn easier actions from the expert that gradually learns more difficult actions. 
\textit{Aggregate Values to Imitate}~\cite{ross2014reinforcement} (AggreVaTe) extends DAgger to minimize \textit{cost-to-go} of a provided expert policy instead of minimizing classification error of replicating the actions, as in the case for DAgger and Coaching. 
The technique is also extended to the reinforcement learning setting, providing strong guarantees for online approximate policy iteration methods. 
This allows a model to reason about how much cost is associated with taking a particular action in the future when subsequent actions can be from the expert policy (i.e cost-to-go). 
In contrast, SEARN uses stochastic policies and rollouts, which can be expensive. 
In contrast, AggreVaTe initially chooses a random time-step $t \in T$ from the expert policy, explores an action and observes the cost-to-go for the expert after performing the action. 
AggreVaTe then trains a model to minimize the expected cost from the generated cost-weighted samples. 
Repeating this process produces a set of cost-weighted training samples that are combined with the original dataset. 
The policy from the learner is then taken up to time $t$ at random, followed by an action, where the expert policy is reintroduced from $t \to T$.
~\newcite{sun2017deeply} extended AggreVaTe to differentiable policy gradient, which leads to faster and better performances, while requiring less training data. 
Both theoretically and empirically they show that the sample complexity for learning is exponentially lower when learning via imitation learning over reinforcement learning-based algorithms. 
Similar to our work, they too provide an option to allow the algorithm to continue executing the expert’s actions with a small probability, instead of always executing $\hat{\pi}_n$ up to a random time-stamp $t$. 
In contrast to our work, this probability is controlled by a dynamic schedule, which samples the experts neighbors, allowing for more exploration throughout the training time than if we had used only the expert policy.
\textit{Locally Optimal Learning To Search} (LOLS)~\cite{chang2015learning} aims to improve upon sub-optimal policies, while providing a local optimality guarantee in relation to the \textit{regret} on the expert policy by competing with one-step deviations from the currently learned policy $\hat{\pi}$. 
LOLS carries this out by minimizing a combination of regret to the reference policy and regret to one-step deviation as $\frac{1}{2}\big(L(\hat{\pi}) - L(\pi^{r})\big) + \frac{1}{2}\big(L(\hat{\pi}) - L(\pi^{d}) \big)$, where the regret is measured as the difference in loss $L$ between the learned policy $\hat{\pi}$ and reference policy $\pi^{r}$ (i.e teacher policy) and $\pi^{d}$ respectively.

\section{Visual Description of NNRS}\label{sec:method}

\begin{figure}
\centering
\captionsetup{justification=centering}
 \includegraphics[scale=.8]{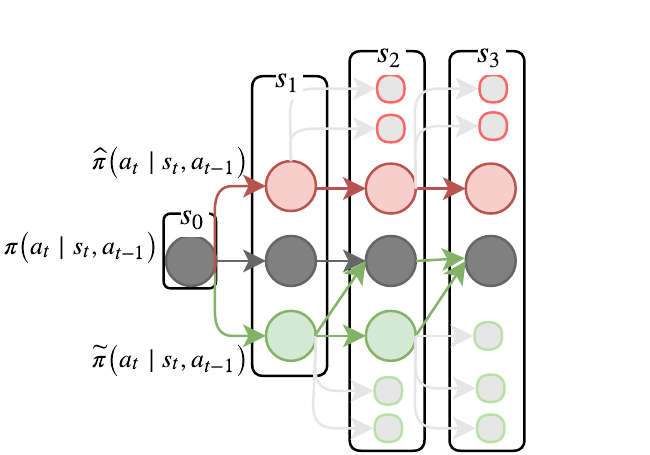}
 \caption{Interpolating between teacher (black), predictions (red) and teacher's neighbors (green)\iffalse teacher's neighbors (green) for target sequence $Y$\fi}\label{fig:stochastic_policy}
\end{figure}

\autoref{fig:stochastic_policy} illustrates the trajectories the sequence might take under a given policy from an initial state $s_0$. 
We show an example where past predictions can lead to a divergence from the teacher trajectory. This can be due to a \textit{difficult to learn} transition from $s_0 \to s_1$, but given a synonymous neighbor (green) as the local one-step deviation allows us to better guide the learner, thus making the transition from $s_1 \to s_2$ easier. We expect the output to be less sensitive to perturbations in the input since the input is locally bounded by the space occupied by the $k$ nearest neighbors of the target $\tilde{y}_{t}$. Likewise, $\tilde{y}_{t}$ can be considered as emulating the problem of compounding errors, since the conditional probability $p(y_{t+1}|x_{1:t},\tilde{y}_{t};\theta)$ is conditioned on the sampled neighbor $\tilde{y}_{t}$ instead of the true target $y_{t}$.

In cases where the model finds it difficult to transition from using $y_{t-1} \to \hat{y}_{t-1}$, interpolating with the neighborhood samples $\tilde{y}$ can provide a smoother policy ($y \to \tilde{y} \to \hat{y}$). Hence, the method can be considered as smoothing which assigns some mass to unseen transitions (similar to Laplacian smoothing) except it is bounded by $k$ neighbors, which is directly proportional to transition probabilities. 
We now consider curriculum schedules to monotonically increase both $\gamma$ and $\gamma_{nnrs}$, corresponding to the sampling rates for SS and NNRS respectively and aim to identify schedules that help mitigate compounding errors by controlling the amount of  exploration of neighbors throughout training.

\section{$k$-NN Sampling Curriculum Schedules}
We consider the neighbors of words as a $3^{rd}$ alternative that helps the teacher guide the learner. In our experiments, we test a linear decay, exponential decay and sigmoid decay, for sampling both target neighbors and predictions sampling introducing some sampling functions such as that shown in \eqref{eq:curs1}, \eqref{eq:curs2} and \eqref{eq:curs3}. In our case, $z = i/\Gamma$ where $\Gamma$ is the total number of training epochs.

\begin{gather}\label{eq:curs1}
g_1(z) = 1 - x/e^{-x} \\ \label{eq:curs2}
g_2(z) = 1/2 + \sin (x\pi - \pi/2) \\ \label{eq:curs3}
g_3(z) = 2/(e^{10x}+1) 
\end{gather}

\begin{figure}[ht]
\begin{center}
 \includegraphics[scale=0.45]{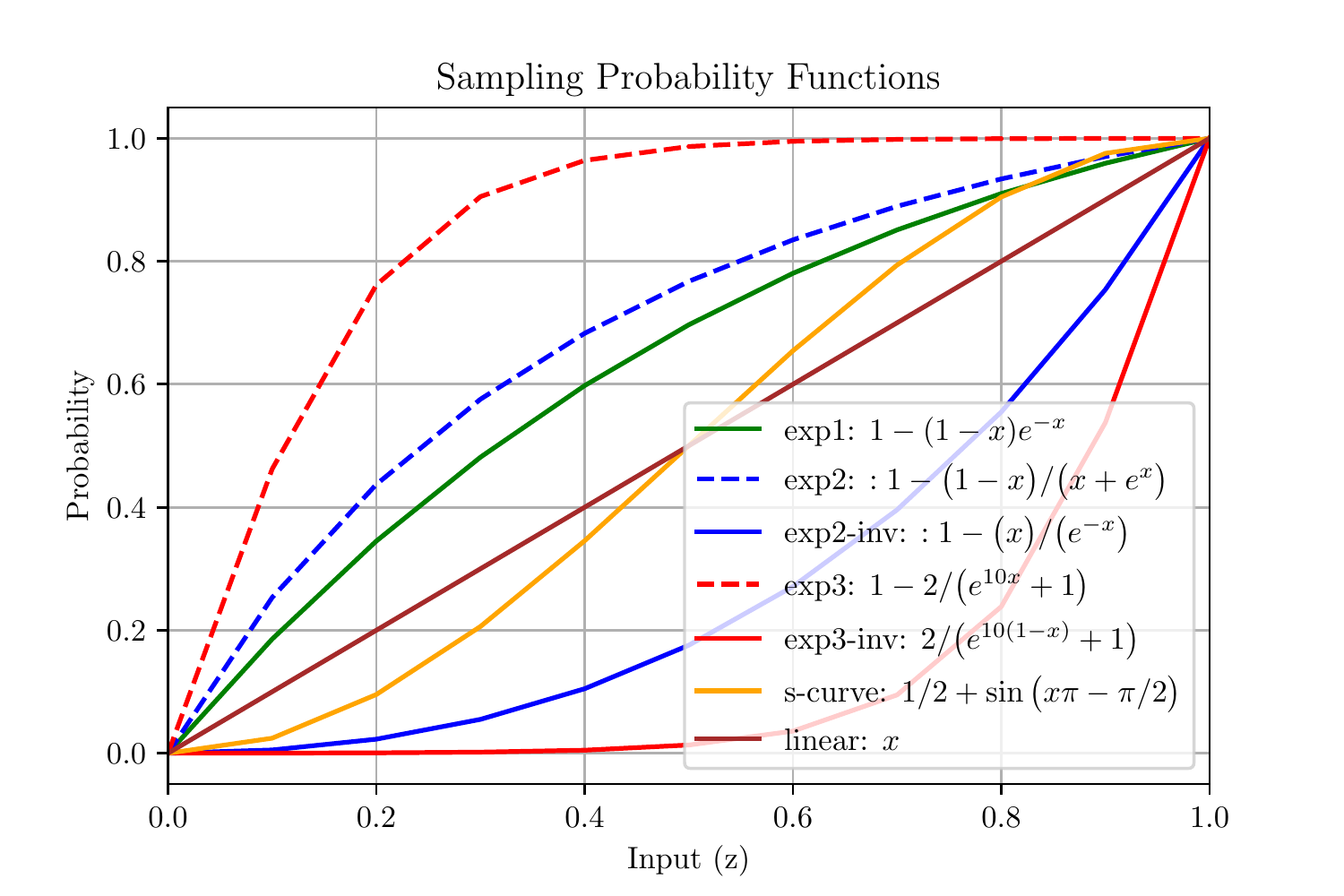}
 \caption{Schedules for sampling rate (x-axis represents scaled $\Gamma$)}\label{fig:spf}
\end{center}
\end{figure}

\section{Connections To $k$-NN Sampling}
Imitation learning algorithms such as DAgger grow the dataset size as new trajectories are created for each policy. Our proposed NNRS method is easily applied online during training and only requires a neighborhood sampling step which is fast and parallelizable. To avoid \textit{snooping}, NNRS does not sample neighbors from outside the vocabulary, only the nearest neighbors within the vocabulary. In comparison to LOLS, we do not require a separate actor network to guide the learner (i.e actor) but rather rely on a curriculum learning based stochastic policy where the expected return is given by the validation perplexity score which is accounted for by adjusting temperature $\tau$ throughout training. As an exploration strategy, $\tau$ is increased throughout training so that neighbors that are further away are gradually assigned a higher sampling probability. LOLS optimizes on the regret to the teacher policy and its own one step deviations. In contrast, this approach optimizes between an interpolation of the regret to the teacher policy, regret to teachers one step deviations and regret to teacher policy with multi-step ahead predictions. In contrast, NNRS ensures the model is robust to local $k$-neighbor prediction errors.

\section{Training Details}
We use a standard 2-hidden layer LSTM sequence model ~\cite{sundermeyer2012lstm} as the basis of our analysis and purposefully avoid using any additional mechanisms that may further obfuscate directly evaluating the effect of sampling strategies for NLM. We test different configurations for sampling strategies, namely: an LSTM with SS with varying $\epsilon$, LSTM with NNRS with varying $\gamma$ and an LSTM with both NNRS and SS. The main baseline is the same LSTM network with no sampling and an LSTM that uses NNRS where the top $k$ neighbors are chosen directly from the transition probability matrix (denoted as TPRS). The former is the main baseline, as the focus of the paper is to identify such sampling techniques improve over the standard maximum likelihood (ML) training. The latter baseline further establishes if creating top $k$ neighbors based on cosine similarity of embeddings is a good approximate to replacement based on the full transition probability matrix. Note that, we use GoogleNews pretrained skipgram~\cite{mikolov2013distributed} embeddings for the training vocabulary embeddings in all subsequent experiments.
For SS and NNRS we consider a range of functions that represent the curriculum learning behavior during training as illustrated in \autoref{fig:spf}. This Figure shows the cumulative distribution for each function, however this assumes a starting NNRS and SS probability $\epsilon_s=0, \gamma_s=0$ and end probability $\epsilon_e=1, \gamma_e =1$ respectively. In experimentation, we report various start and end ranges for both $\epsilon$ and $\gamma$ with these functions.


\begin{figure}
\centering
\captionsetup{justification=centering}
 \includegraphics[scale=0.5]{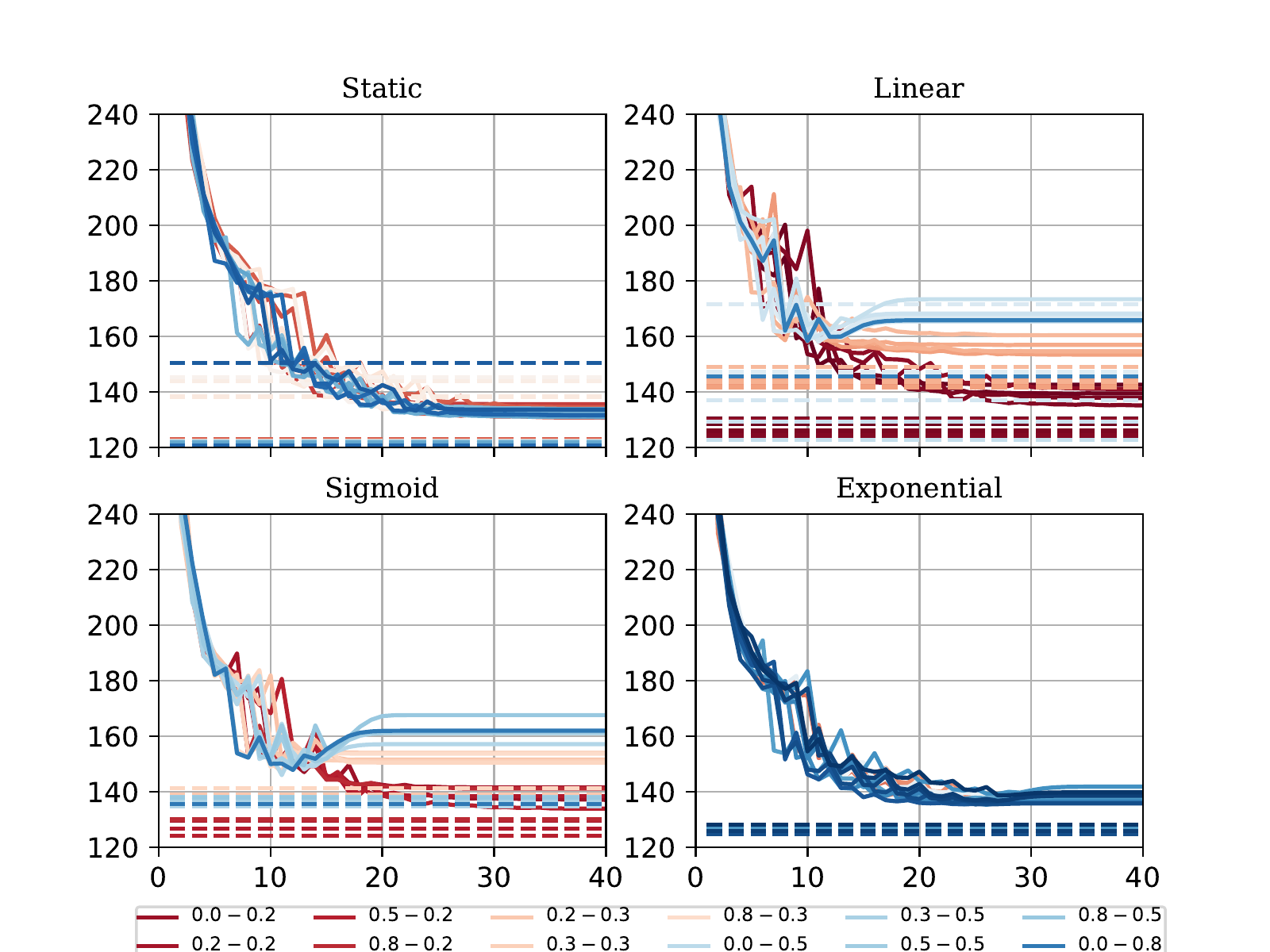}
 \caption{Perplexity on WikiText2: SS and NNRS (best viewed in color)}\label{fig:wiki2_lc_ss}
\end{figure}

\subsection{Additional PTB Experimental Results}

\begin{figure}
\centering
\captionsetup{justification=centering}
 \includegraphics[scale=0.5]{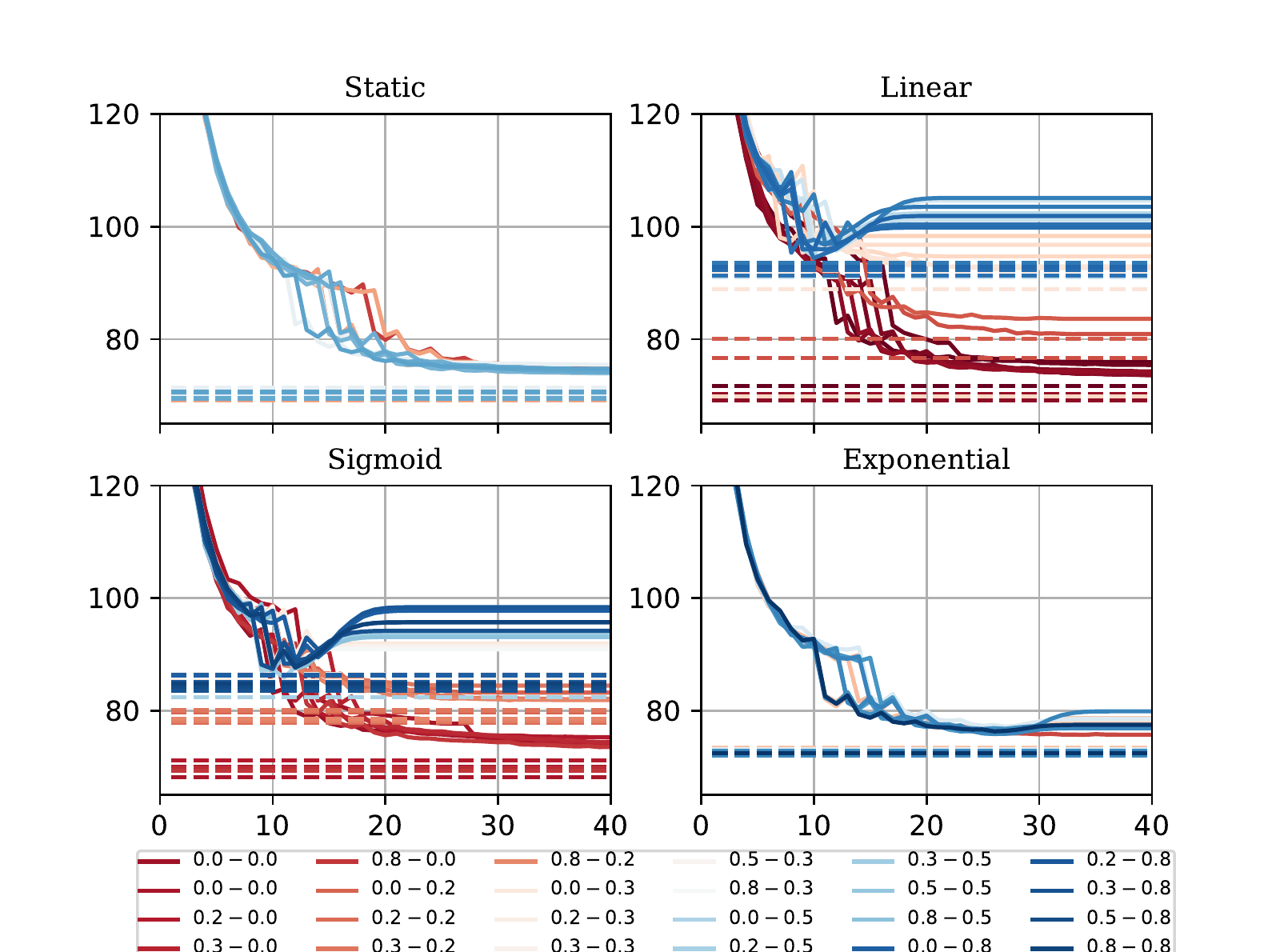}
 \caption{Perplexity on Penn-Treebank}\label{fig:ptb_lc_ss}
\end{figure}

\paragraph{Penn-Treebank Performance}
\autoref{fig:ptb_lc_ss} shows the results of the proposed approach on Penn-Treebank. For PTB, we find that static and exponential settings for both SS and NNRS show best performance and converge quicker. In all cases, convergence for all functions behaves as expected for $\{\sigma, \gamma\} < 0.3$. We also see the same trend as with WikiText-2, that exponential function allows for higher sampling rates when compared to linear and sigmoid functions. This further suggests that NNRS is most effective near convergence, as the sampling probability exponentially increases while the validation perplexity begins to plateau.

\bibliographystyle{acl_natbib}
\bibliography{acl2020}